\documentclass{article}
\usepackage{xcolor}
\usepackage{xspace}
\definecolor{olive}{rgb}{0.5,0.5,0}
\usepackage[preprint]{corl_2026} % Uncomment for pre-prints (e.g., arxiv); This is like ``final'', but will remove the CORL footnote.
\usepackage{alias}
\usepackage{graphicx}
\usepackage{subcaption}
\usepackage{wrapfig}
\usepackage{booktabs}
\usepackage{multirow}
\usepackage{etoc}
\usepackage{placeins}
\usepackage{titletoc}
\usepackage{cuted}
\title{\ourmethod{}: \underline{A}ligning \underline{R}obot Scene \underline{G}eometry \underline{U}nder \underline{S}hifting Views with Large 3D Vision Models}

\author{
  Rishik Sathua\textsuperscript{1} \quad Haonan Chen\textsuperscript{2} \quad Katherine Driggs-Campbell\textsuperscript{1}\\
  \textsuperscript{1}University of Illinois at Urbana-Champaign \quad \textsuperscript{2}Harvard University\\
  \texttt{\url{https://rsathua.github.io/ARGUS/}} \\
}

\begin{document}
\maketitle

%===============================================================================

\begin{abstract}
    Large-scale visuomotor policies have demonstrated impressive performance across a wide range of robot manipulation tasks. However, despite this success, manipulation polices often entangle scene geometry with the corresponding viewpoint, learning where objects lie in an image rather than where it lies in the task space. This entanglement inherently limits the corresponding policy's ability to learn from viewpoint-diverse datasets (ex. DROID, BridgeV2) and generalize beyond the viewpoints captured in their training data. In this work, we present \ourmethod{}, an observation pre-processing pipeline that uses large-scale 3D vision models to align image observations from arbitrary camera viewpoints into a canonical viewpoint before passing it to downstream visuomotor policies. Experiments across training datasets with varying levels of viewpoint diversity, from fixed multi-view camera configurations to highly varied camera placements, show that our method consistently outperforms prior approaches across both limited-view and view-diverse training regimes. In efficiency comparisons, \ourmethod{} demonstrates an ability to learn from view-diverse data, converging to high success rates 4–6× faster than previous methods by leveraging a simplified observation space. Overall, our findings show that leveraging large-scale 3D vision models reduces the learning burden on visuomotor policies, enabling more efficient learning from large-scale, viewpoint-diverse robot datasets.
    
\end{abstract}

\section{Introduction}
A core challenge in visuomotor learning stems from the entanglement of task-relevant information with scene-specific variation. Identical tasks involving the same objects may yield drastically different visual representations under changes in viewpoint, embodiment, or environment, despite sharing the same underlying semantics. This variability makes learning task-centric representations difficult and brittle, ultimately limiting generalization to novel scene configurations. One common approach to improving robustness is to train policies on large and diverse datasets that encompass such variations~\cite{khazatsky2024droid, open_x_embodiment_rt_x_2023, walke2023bridgedata, jang2021bc}. While this strategy has shown success~\cite{black2026pi0visionlanguageactionflowmodel, octo_2023}, we show in Section~\ref{sec:efficiency_eval} that forcing models to disentangle task semantics from scene-induced variability can be data-inefficient and costly. This is consistent with prior work~\cite{xie2023decomposinggeneralizationgapimitation, pumacay2024colosseumbenchmarkevaluatinggeneralization, gao2024efficientdatacollectionrobotic}, which finds that RGB-based policies trained under fixed viewpoints often fail to generalize to new camera configurations, while those trained on datasets collected under unconstrained viewpoints also struggle to generalize effectively to novel viewpoints.

In this paper, we propose an alternative approach. Instead of fitting policies to highly diverse raw observations in the hope that invariances will emerge, we simplify the visual observation space using pretrained vision models that capture rich 3D scene structure. Specifically, given a dataset with freely moving cameras (subject to always observing the task-relevant workspace), we reconstruct a canonicalized view by lifting observations into a 3D representation using a pretrained 3D vision model and re-rendering them from a fixed reference viewpoint before passing them to downstream visuomotor policies. We refer to this approach as \textbf{Aligning Robot scene Geometry Under Shifting views (ARGUS)}.

\ourmethod{} offers several advantages. First, although we evaluate it in the context of imitation learning, it is a preprocessing pipeline and can be integrated into a wide range of visuomotor learning paradigms, enabling more efficient learning and improved generalization from training datasets. Second, because it leverages pretrained 3D vision models to reconstruct scene geometry, \ourmethod{} is naturally complementary to advances in 3D perception and is expected to benefit from continued progress in this area. Finally, by relying on learned 3D reconstruction rather than explicit depth sensors, \ourmethod{} reduces the need for specialized hardware, lowering the barrier to achieving viewpoint robustness in real-world settings.

We highlite the main contributions of this paper below:
\begin{itemize}
\item We present \ourmethod{}, a general-purpose viewpoint normalization pipeline that reduces viewpoint-induced scene variation by aligning observations to a canonical reference frame, enabling more effective policy learning.

\item We evaluate \ourmethod{} under both diverse-viewpoint and single-viewpoint dataset settings, demonstrating data efficiency and strong zero-shot generalization to unseen camera viewpoints.

\item We compare \ourmethod{} against depth-based methods and show competitive performance, suggesting that learned 3D reconstruction can serve as an effective alternative to explicit depth sensing for viewpoint-robust visuomotor learning.

\end{itemize}

\section{Related Works}
\subsection{Dataset Synthesis/Augmentation for Viewpoint Invariance}
Learning viewpoint-invariant policies has long been a popular goal in robot learning. A common approach is dataset synthesis, where multi-view data is generated in simulation to improve robustness to camera variation~\cite{seo2023multiviewmaskedworldmodels, sadeghi2017sim2realviewinvariantvisual}. While effective for improving viewpoint generalization, these methods must contend with the additional challenges of sim-to-real transfer inherent to visual simulation pipelines. Another line of work performs real-world dataset augmentation using view augmentation methods~\cite{tian2024vista, chen2025tool}, thereby avoiding sim-to-real issues. However, these approaches can introduce visual artifacts from the synthesis process and often degrade in performance under large viewpoint shifts. As a result, evaluations are frequently limited to small perturbations around the training viewpoints. In contrast, grounding observations in explicit 3D structure enables more principled extrapolation to large viewpoint changes. In our experiments, we evaluate robustness under viewpoint shifts of up to $135^\circ$ (Section~\ref{sec:Viewpoint_Variance_Evaluation}), covering a wide range of camera perspectives to assess viewpoint robustness.

\subsection{Camera Ray Encoding with Compositional Data Collection}

Camera ray encoding is a widely used technique in the vision literature for modeling 3D scene structure from images~\cite{gao2024cat3d,mildenhall2020nerfrepresentingscenesneural,zhang2024camerasraysposeestimation}. This idea has also been adopted in robot learning, where camera rays are encoded as Pl\"ucker coordinates within visuomotor policies~\cite{jiang2025knowcameraisviewinvariant}. Beyond ray encoding, \cite{jiang2025knowcameraisviewinvariant} further encourages viewpoint invariance and data efficiency through compositional data collection strategies~\cite{xie2023decomposinggeneralizationgapimitation}, where consecutive demonstrations share overlapping viewpoints. This structure enables models to more effectively disentangle viewpoint variation from scene content during training. However, such compositional assumptions are difficult to guarantee in large-scale, real-world datasets~\cite{khazatsky2024droid, walke2023bridgedata}, where camera viewpoints are typically uncontrolled. In contrast, \ourmethod{} does not rely on any assumptions about viewpoint structure in the dataset, while still enabling efficient and generalizable learning.

\subsection{3D Scene Modeling with Depth}
Depth-based policies often exhibit improved viewpoint robustness, as their representations are grounded in 3D structure that is less sensitive to camera pose. More broadly, structured intermediate representations have been used to improve manipulation generalization by decoupling task-relevant structure from low-level action generation, including through behavior primitives and predicted object interactions~\cite{chen2023predicting}. One line of work directly learns from 3D representations such as point clouds, voxel grids, or feature fields~\cite{shridhar2022peract, grotz2024peract2benchmarkinglearningrobotic, gervet2023act3d3dfeaturefield, ze2024gnfactormultitaskrealrobot, gkanatsios20253dflowmatchactorunified,3d_diffuser_actor}. Another line of work constructs explicit 3D scene representations, which are then rendered into canonical viewpoints for downstream policy learning~\cite{goyal2023rvt, goyal2024rvt2learningprecisemanipulation}. While effective, these approaches often rely on depth sensors to obtain 3D structure. While they are generally effective, depth sensors are less widely available than RGB cameras, and their performance can vary significantly across hardware platforms~\cite{Rustler_2025}, posing challenges for scalable and consistent deployment. Motivated by these limitations, we focus on RGB-based sensing, which is more ubiquitous and provides more consistent and reliable performance across hardware platforms.

\section{Approach}
\label{sec:method}
\subsection{Preliminaries}
While formulating \ourmethod{}, we assume access to an imitation learning dataset with at least two calibrated external cameras. Since our study focuses on viewpoint generalization, we hold the task scene fixed between training and evaluation; generalization across scenes is orthogonal to our contribution and outside the scope of this work. We further assume that the cameras, regardless of their general position in the world, are always pointed towards the task space. Beyond these assumptions, the camera viewpoints are unrestricted and can be visualized in Figure~\ref{fig:cam_poses}.

\subsection{Overview}
\begin{figure}[t]
    \centering
    \includegraphics[width=\linewidth]{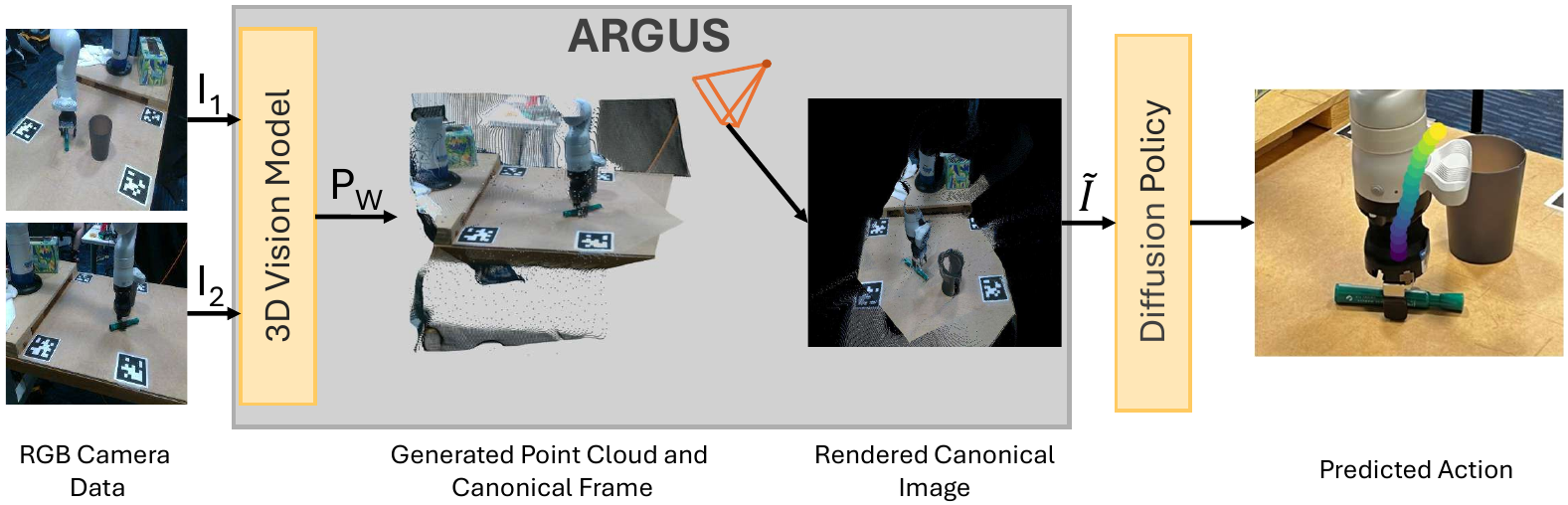}
    \caption{\textbf{Overview of \ourmethod{}.} Given two RGB images from any two camera viewpoints, we first construct a 3D point cloud of the scene using a Large Vision Model (ex. VGGT~\cite{wang2025vggt}). The point cloud is then aligned and scaled to fit the approximate geometry of the real world before it is used to render a 2D RGB image from a canonical viewpoint. This rendered image is then fed into the downstream visuomotor (ex. Diffusion) policy which is then used to predict actions.}
    \label{fig:summary}
\end{figure}
\ourmethod{} reformulates visuomotor policy learning as a two stage problem. Rather than learning a direct mapping from viewpoint-varying observations $I$ to actions $A$, we decompose the problem into first normalizing the viewpoint with $h:I\rightarrow \tilde{I}$ and then map the normalized viewpoint observation to an action using a downstream visuomotor policy $f_\theta:\tilde{I}\rightarrow A$. Within this two stage decomposition, \ourmethod{} plays the role of $h$, a function that normalizes images from unconstrained viewpoints.

As seen in Figure~\ref{fig:summary}, \ourmethod{} can be broken down into three steps in which we (i) estimate a scene point cloud with a large 3D vision model, (ii) recover its metric scale and align it to the robot base frame using the known real-world camera extrinsics, and (iii) render the aligned point cloud from a fixed canonical viewpoint. The resulting canonical image is the input to the policy at both training and inference time, which removes viewpoint as a nuisance variable from the policy's learning problem.

\subsection{Cross-View Point Cloud Estimation}
\begin{wrapfigure}{r}{0.5\textwidth}
    % \vspace{-2mm}
    \centering
    \includegraphics[width=0.5\textwidth]{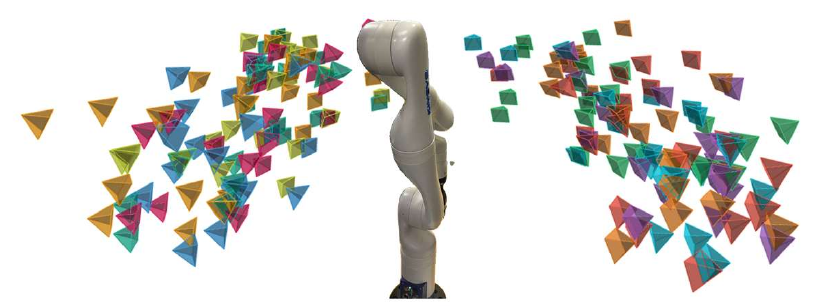}
    \caption{\textbf{\boldmath$D_{Diverse}$ Viewpoints.} Visualization of the camera pose diversity present within our diverse training datasets.}
    \label{fig:cam_poses}
    % \vspace{-3mm}
\end{wrapfigure}
To unify the diverse visual representations from unconstrained camera viewpoints in our datasets, we take inspiration from previous works in depth-based robot manipulation policies~\cite{3d_diffuser_actor, goyal2023rvt, gkanatsios20253dflowmatchactorunified} and construct a point cloud of the scene. However, since our goal is to generalize across viewpoints without relying on the availability of depth information, we compile such point clouds using an off-the-shelf large 3D vision model, VGGT~\cite{wang2025vggt}, that takes as input images from multiple viewpoints and outputs a predicted point cloud and corresponding predicted extrinsics. It is important to note that the point cloud and extrinsics VGGT outputs are not accurate in scale and are constructed with respect to one of the input camera's frames.

\subsection{Metric Scaling and World-Frame Alignment}
\label{sec:alignment_sec}
Because VGGT's predicted geometry is defined only up to an unknown global scale, we recover the metric scale from the ratio between the inter-camera baseline in the real-world extrinsics and the corresponding baseline in the VGGT-estimated extrinsics. We then transform the scaled point cloud into the robot base frame using the known real-world camera extrinsics. This two-step scaling and alignment is given by Equation~\ref{eq:pointcloud_transform}.
\begin{equation}
\mathbf{P}_A=\frac{t_A-t_B}{t^{VGGT}_A-t^{VGGT}_B}\mathbf{P}^{VGGT}_A
\qquad
\mathbf{P}_W
= 
{}_{W}\mathbf{T}_{A}
\mathbf{P}_A
\label{eq:pointcloud_transform}
\end{equation}
Here, $\mathbf{P}^{VGGT}_A$ and $\mathbf{P}_A$ denote the VGGT-predicted and scaled point clouds in camera $A$'s coordinate frame respectively, $\mathbf{P}_W$ is the scaled point cloud expressed in the world frame, ${}_{W}\mathbf{T}_{A}$ is the extrinsic matrix of camera $A$ representing its pose with respect to the world frame $W$, $t_A$ and $t_B$ denote the translation components of camera $A$ and $B$'s real-world extrinsic matrices, and $t^{VGGT}_A$ and $t^{VGGT}_B$ denote the corresponding translation components in the VGGT-estimated extrinsic matrices. Cameras $A$ and $B$ refer to any two of the available calibrated cameras; when more than two cameras are present, we use the pair with the largest real-world baseline for numerical stability.

\begin{wrapfigure}{r}
{0.5\textwidth}
    \vspace{-30pt}
    \centering
    \includegraphics[width=0.215\textwidth]{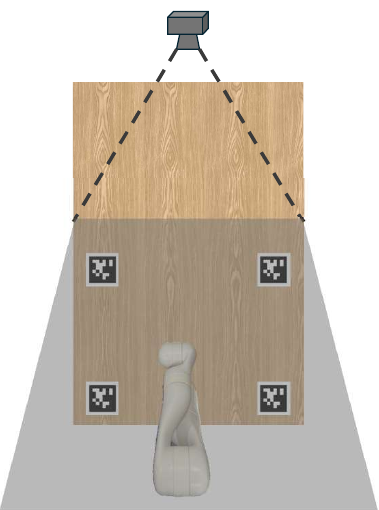}
    \includegraphics[width=0.275\textwidth]{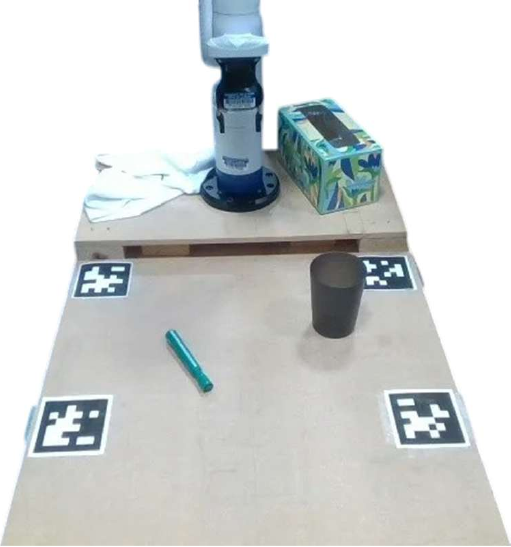}
    \hfill
    
    \caption{\textbf{Canonical Viewpoint Selection}.
The canonical viewpoint is chosen to maximize visibility of the workspace and manipulated objects.}
    \vspace{-17pt}
    \label{fig:canonical_selection}
    
\end{wrapfigure}
\subsection{Canonical Viewpoint Rendering}
Given the scaled and world-aligned point cloud $\mathbf{P}^{scaled}_W$, we render the scene from a fixed virtual canonical viewpoint expressed in the robot base frame. This canonical pose is chosen to maximize visibility of the workspace (as seen in Figure~\ref{fig:canonical_selection}), and is held constant across training and evaluation so the downstream policy always observes the scene from the same vantage point regardless of where the input cameras were placed. The rendered canonical image is then passed to the downstream visuomotor policy that controls the robot's behavior.

\section{Experiments}
\label{sec:experiments}
\begin{figure*}[t]
    \centering
    \begin{subfigure}{0.25\textwidth}
        \centering
        \includegraphics[width=\linewidth]{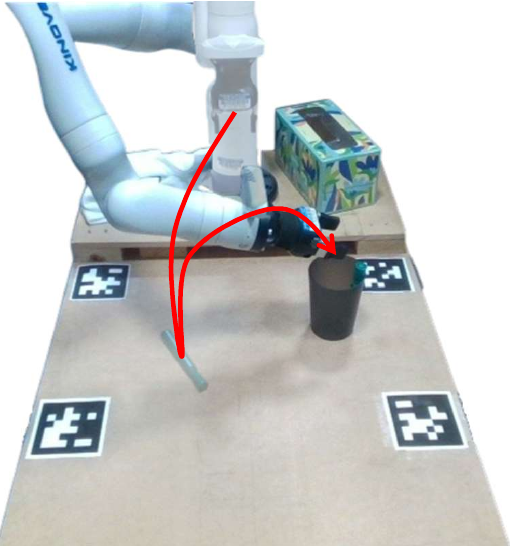}
        \subcaption{Put Marker In Cup}
    \end{subfigure}%
    \begin{subfigure}{0.25\textwidth}
        \centering
        \includegraphics[width=\linewidth]{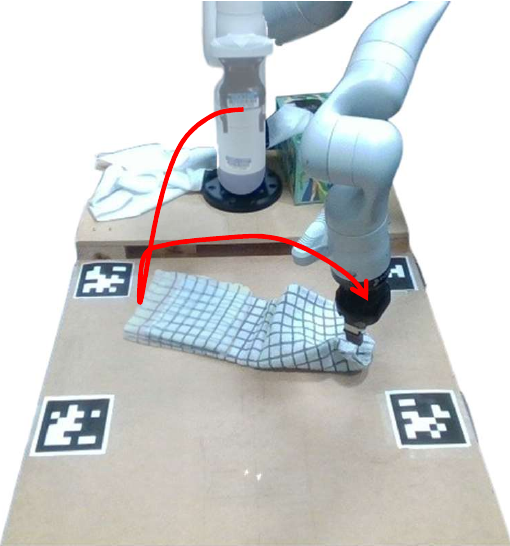}
        \subcaption{Unfold Towel}
    \end{subfigure}%
    \begin{subfigure}{0.25\textwidth}
        \centering
        \includegraphics[width=\linewidth]{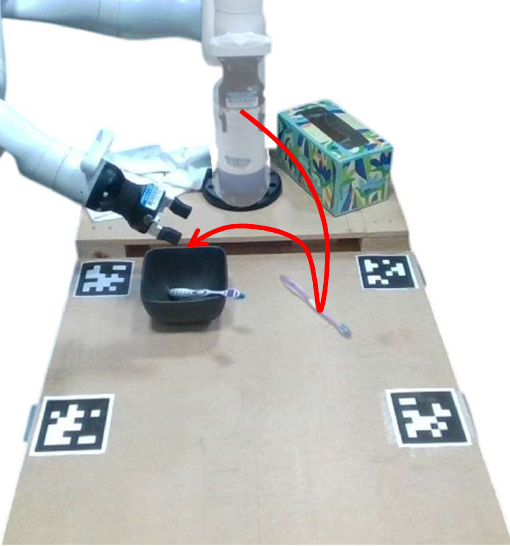}
        \subcaption{Put Brush In Bowl}
    \end{subfigure}%
    \begin{subfigure}{0.25\textwidth}
        \centering
        \includegraphics[width=\linewidth]{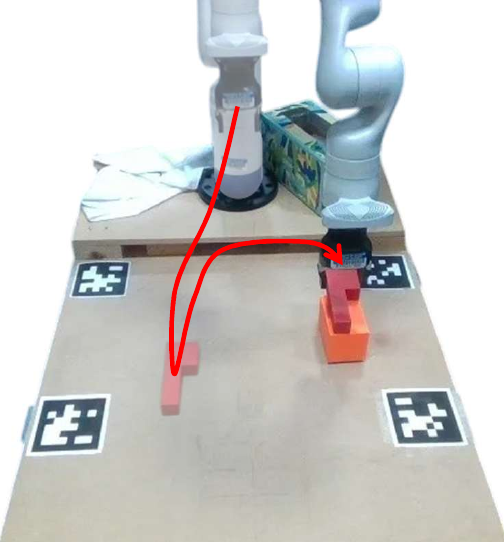}
        \subcaption{Stack Blocks}
    \end{subfigure}
    \caption{\textbf{EvaluationTasks.} Real-world evaluation across four manipulation tasks.}
    \label{fig:tasks}
\end{figure*}
To study \ourmethod{}'s effectiveness as a data pre-processing pipeline, we evaluate it through numerous experiments each exploring a different question relevant to effectiveness, efficiency, and generalizability. Specifically, we ask the following questions:
\begin{itemize}
    \item Does \ourmethod{} help robot manipulation policies to learn from datasets with varying levels of viewpoint diversity?
    \item How does \ourmethod{} impact robot policies' training time and sample efficiency when tasked to fit complex and viewpoint diverse datasets?
    \item How does \ourmethod{} compare to depth-based methods that are naturally robust to viewpoint variation?
\end{itemize}
We evaluate these questions through four tasks in the real world: 2 pick and place, 1 stacking, and 1 unfolding tasks. These tasks can be viewed in Figure~\ref{fig:tasks}. We select tasks involving relatively small objects that can be sampled across the workspace, enabling systematic evaluation of whether the policy can robustly spatially reason about and perform the task efficiently.

\subsection{High/low Viewpoint Variance Evaluation}
\label{sec:Viewpoint_Variance_Evaluation}
\subsubsection{Experimental Setup}

\begin{figure*}[t]
    \centering
    \begin{subfigure}{0.17\textwidth}
        \centering
        \includegraphics[width=\linewidth]{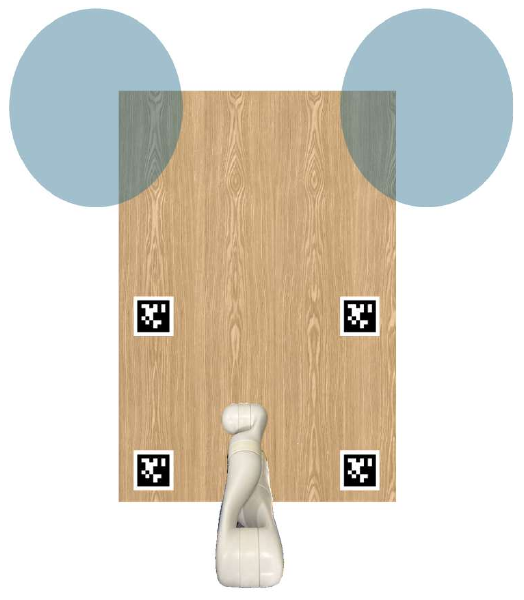}
        \subcaption{}
    \end{subfigure}%
    \begin{subfigure}{0.19\textwidth}
        \centering
        \includegraphics[width=\linewidth]{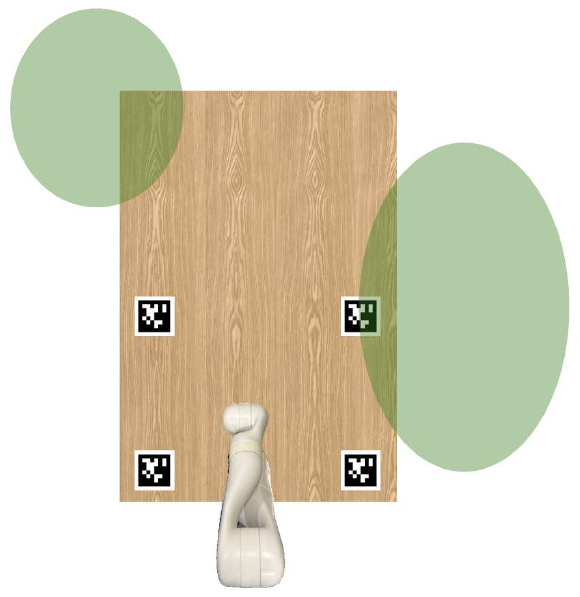}
        \subcaption{}
    \end{subfigure}%
    \begin{subfigure}{0.20\textwidth}
        \centering
        \includegraphics[width=\linewidth]{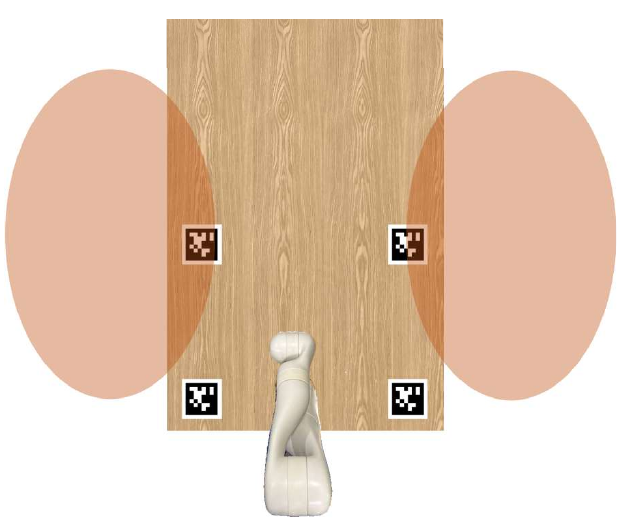}
        \subcaption{}
    \end{subfigure}%
    \begin{subfigure}{0.18\textwidth}
        \centering
        \includegraphics[width=\linewidth]{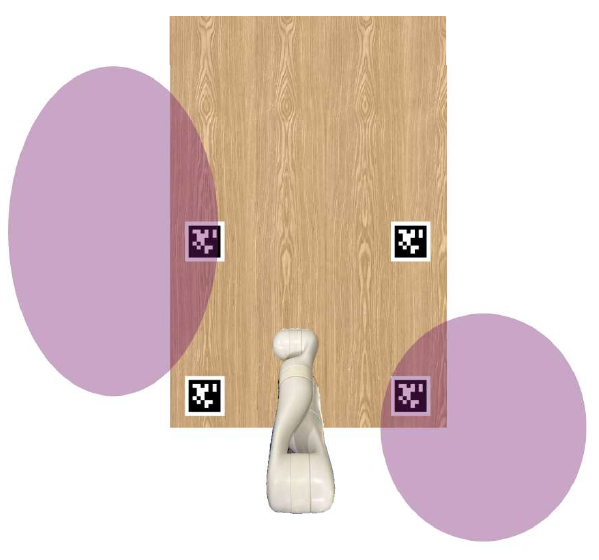}
        \subcaption{}
    \end{subfigure}%
    \begin{subfigure}{0.171\textwidth}
        \centering
        \includegraphics[width=\linewidth]{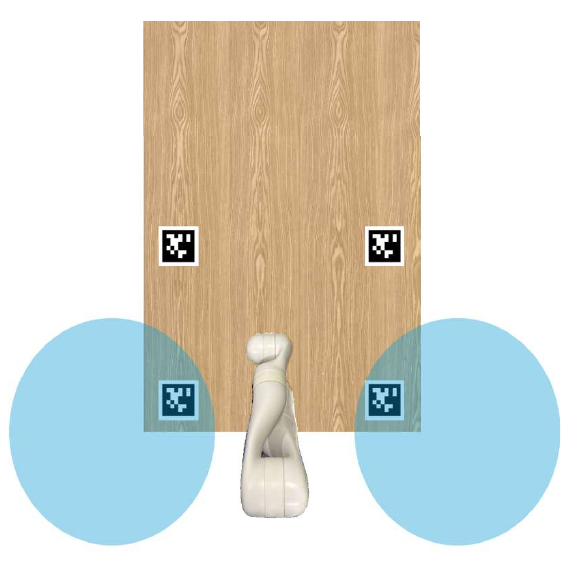}
        \subcaption{}
    \end{subfigure}
    \caption{\textbf{Evaluation-time Camera Viewpoint Positioning.} Top-down visualization of camera sampling distributions across five configuration groups. Ellipses indicate the spatial extent of camera positions within each group.}
    \label{fig:eval-cams}
\end{figure*}

We study \ourmethod{}'s performance with two different types of training datasets, one with limited, fixed viewpoints and one with highly diverse viewpoint positions.  
Evaluating \ourmethod{} under both these settings allows us to confirm viability in different data collection paradigms and to study two important capabilities: (1) the ability to generalize to novel viewpoints despite having no viewpoint variation at training time and (2) the ability to extract useful information from highly diverse and unstructured viewpoint datasets. 

\begin{wrapfigure}{r}{0.43\textwidth}
    % \vspace{-2mm}
    \centering
        \includegraphics[width=0.4\textwidth]{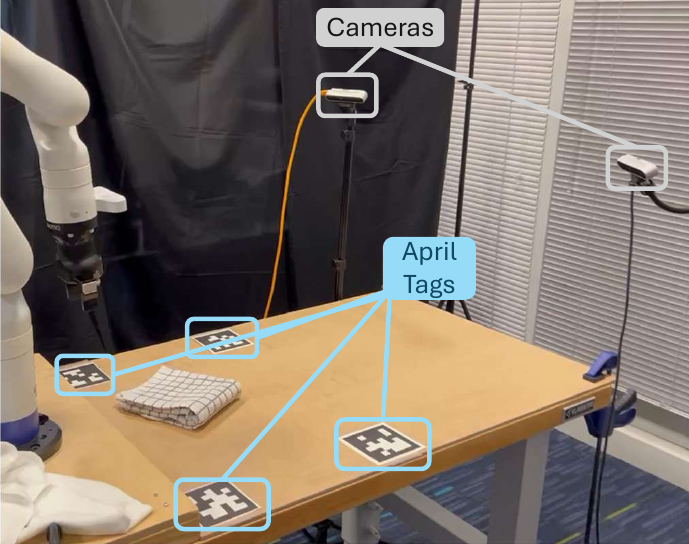}
    \caption{\textbf{Experimental Setup.} Labelled experimental setup highliting variable position cameras and april tags used to calibrate them.}
    \label{fig:setup}
    \vspace{-5mm}
\end{wrapfigure}

The dataset with fixed viewpoints, $D_{fixed}$, was compiled by fixing the cameras' locations at two poses at the head of the table (visualized in Figure~\ref{fig:eval-cams} configuration (a)). 
The diverse dataset, $D_{diverse}$, was constructed by sampling random camera viewpoints after collecting each demo. We collect 100 demos following this protocol accumulating 200 unique camera positions per task. An example of such an instance of $D_{diverse}$ can be found in Figure~\ref{fig:cam_poses}. It is important to note, we use AprilTags~\cite{apriltags} to provide \ourmethod{} and our baselines with camera extrinsic estimates as seen in Figure~\ref{fig:setup}.

In both of these setting, we study the success rates of policies across 5 different camera configurations with each configuration sampled from a region about the table perimeter. These sample regions can be visualized in Figure~\ref{fig:eval-cams} spanning the length of the table. Under each of the 5 camera configurations, we study each method's performance manipulating objects in 10 random object configurations. We report our findings as each policy's average success rate across the 50 rollouts (5 camera configurations $\times$ 10 object configurations) and standard error across the 5 camera configurations.

When training with $D_{diverse}$, we compare \ourmethod{} to KYC~\cite{jiang2025knowcameraisviewinvariant}, a past work that encodes the camera extrinsics as 6D Pl\"ucker coordinates directly into the down stream policy. KYC was specifically designed for learning from datasets with high viewpoint diversity making it a natural comparison in our study. For the $D_{fixed}$ setting, we compare \ourmethod{} to VISTA~\cite{tian2024vista}, which augments training data with fixed viewpoints (ex. $D_{fixed}$) to mimic a more viewpoint-diverse one (ex. $D_{diverse}$) by imagining what the scene would look like from random viewpoints using a novel view synthesis model ZeroNVS~\cite{zeronvs}. This data augmentation attribute that allows it to generalize beyond limited training data makes VISTA well suited as a comparison for the $D_{fixed}$ setting of our study. Since both of these methods natively use diffusion policy as their downstream visuomotor policy, we do the same to ensure direct comparability in both settings. 
\begin{figure}[t]
    \centering
    \begin{subfigure}{0.48\textwidth}
        \centering
        \includegraphics[width=\linewidth]{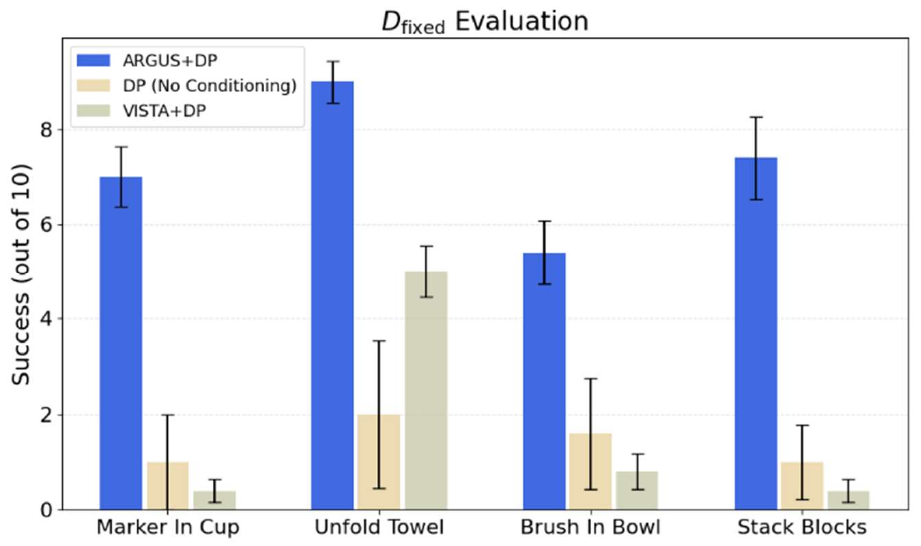}
        \label{fig:training_efficiency}
    \end{subfigure}%
    \hspace{0.01\textwidth}
    \vrule
    \hspace{0.01\textwidth}
    \begin{subfigure}{0.48\textwidth}
        \centering
        \includegraphics[width=\linewidth]{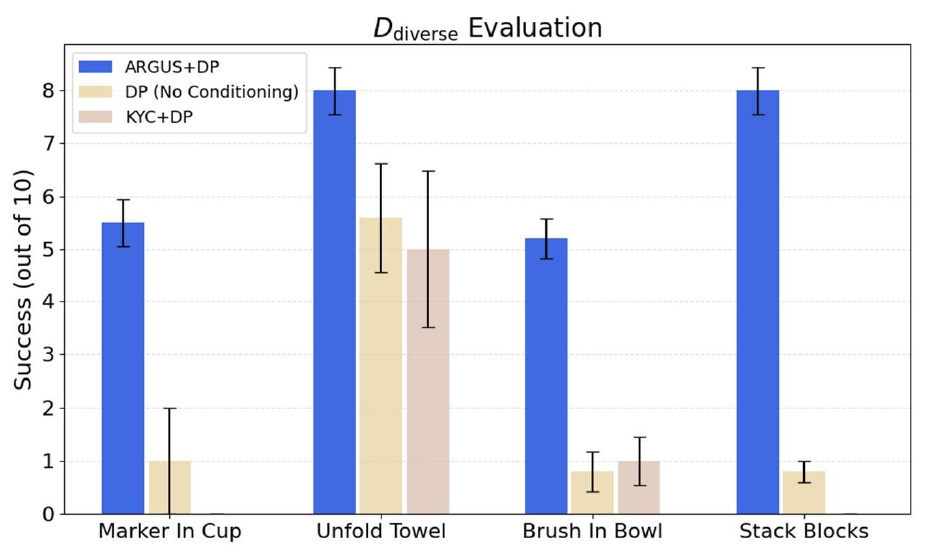}
        \label{fig:data_efficiency}
    \end{subfigure}
    \caption{\textbf{Average success rates across five novel multi-view camera configurations.} (Left) Methods are trained on single-viewpoint configuration datasets and evaluated for generalization to unseen viewpoints. (Right) Methods are trained on viewpoint-diverse datasets and evaluated under the same novel, previously unseen viewpoints. }
    \label{fig:viewpoint_variance_eval}
\end{figure}
Finally, to measure comparative impact of our and past methods on diffusion policy, we also compare against classical diffusion policy with no modifications or conditioning under both settings.

All variants of diffusion policy (\ourmethod{}, KYC, VISTA, and classical) were trained using datasets with 100 demonstrations for 1500 epochs. Diffusion Policies trained with \ourmethod{} and VISTA are trained to predict actions relative to the world frame while KYC is trained to predict relative actions (the best performing action space as per~\cite{jiang2025knowcameraisviewinvariant}'s findings). It is important to note that VISTA~\cite{tian2024vista} was initially designed as a policy with one external camera with novel viewpoint evaluation extending to at most $45^\circ$ beyond the fixed camera pose in $D_{fixed}$. To ensure fairness we give VISTA access to both external cameras used in our evaluation setting (augmenting both images to synthesize ones from random viewpoints). Additionally, since we extend the scope of viewpoint evaluation from $45^\circ$ to roughly $135^\circ$, we follow VISTA's findings and finetune ZeroNVS on $D_{diverse}$ to ensure best possible performance in this increased variation setting. 

\subsubsection{Results}
We report our findings in Figure~\ref{fig:viewpoint_variance_eval} where we find \ourmethod{} outperforms KYC, VISTA, and classic Diffusion Policy by a considerable margin while demonstrating a relatively consistent success rate (observed by smaller standard error bars) across all five camera viewpoint configurations. 
% The results of this study can further be seen broken down by camera configuration in \rs{fix and refer to Appendix here}

\subsection{Training Efficiency Evaluation}
\label{sec:efficiency_eval}
\subsubsection{Experimental Setup}

In this section, we look to understand \ourmethod{}'s impact on the overall training efficiency of its downstream visuomotor policy when tasked to fit the largely diverse dataset $D_{diverse}$ (see~\ref{sec:Viewpoint_Variance_Evaluation} for dataset details). Specifically, we study its impact on how it influences training time (measured in number of dataset passes) and data efficiency (measured in number of demos present in training data). For best interpretability, we conduct this evaluation on the task with the highest average success rate from~\ref{sec:Viewpoint_Variance_Evaluation}, which is the unfolding towel task as seen in Figure~\ref{fig:viewpoint_variance_eval}. For this study, we use a single viewpoint configuration sampled from region (a) (as displayed in Figure~\ref{fig:eval-cams}) and keep it constant throughout the evaluation. 

Since our focus for this study is to evaluate each method's efficiency when fitting an unstructured dataset like $D_{diverse}$, we compare \ourmethod{} to KYC and classical diffusion policy. Except for the training time and dataset, which we vary by the trial, all training parameters are maintained as discussed in~\ref{sec:Viewpoint_Variance_Evaluation}.

\subsubsection{Results}
\begin{figure}[t]
    \centering

    \begin{subfigure}[t]{0.48\textwidth}
        \centering
        \includegraphics[
            width=\linewidth,
        ]{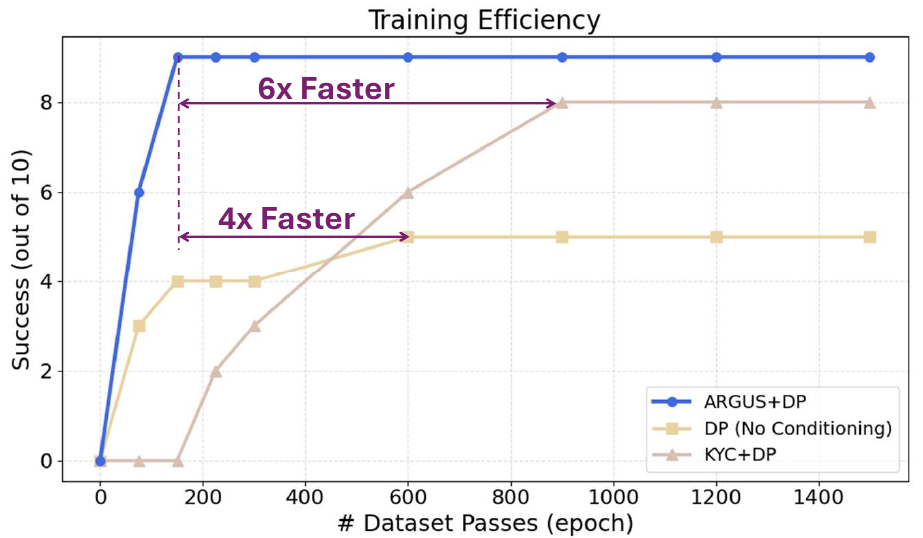}
        \label{fig:training_efficiency}
    \end{subfigure}
    \hspace{0.01\textwidth}
    \vrule
    \hspace{0.01\textwidth}
    \begin{subfigure}[t]{0.48\textwidth}
        \centering
        \includegraphics[
            width=\linewidth,
        ]{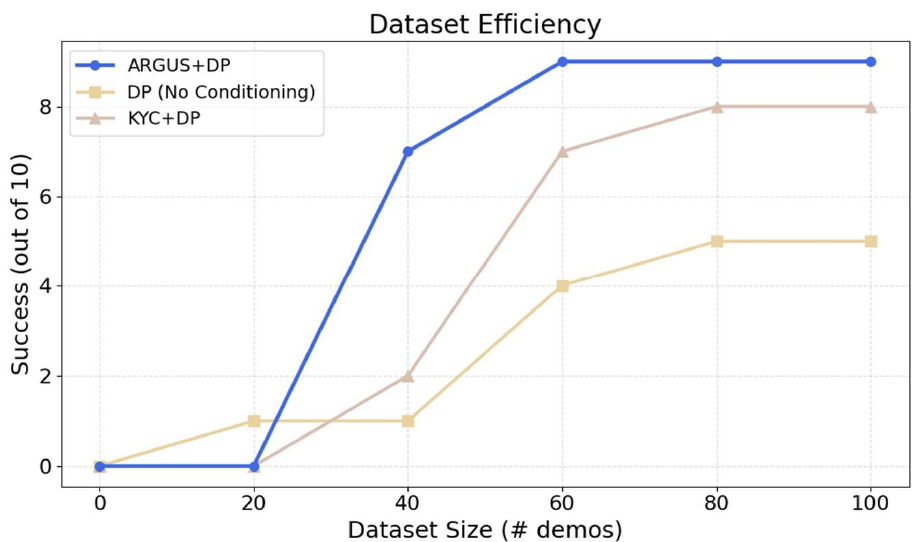}
        \label{fig:data_efficiency}
    \end{subfigure}

    \caption{Training and sample efficiency across methods. (Left) Training efficiency measured by the number of dataset passes required for each method to converge (Right) Sample efficiency, showing performance as a function of number of training samples.}
    \label{fig:efficiency_comparison}
\end{figure}

We report our overall training efficiency findings in Figure~\ref{fig:efficiency_comparison} where we find \ourmethod{} converges at a 90\% success rate 6x faster than KYC and 4x faster than classical diffusion policy. Additionally, we found \ourmethod{} achieves a high (70\%) success rate even when trained with only 40 demonstrations (80 unique viewpoints), indicating strong data efficiency compared to KYC and classical DP. We attribute this boost in efficiency to \ourmethod{} simplifying the data space from unconstrained viewpoints to a controlled canonical ones, thus easing the learning burden off the model and allowing it to learn from structured input despite the unstructured dataset.

\subsection{Comparison to Depth-Based Methods}
\subsubsection{Experimental Setup}

In this portion of our study we seek to understand \ourmethod{}'s performance when applied and compared to depth or point cloud based architectures that are naturally good at viewpoint invariance. For this study, we chose to compare \ourmethod{} with 3D Diffuser Actor~\cite{3d_diffuser_actor}. Specifically, we evaluated how 3D Diffuser Actor's performance changed when we applied \ourmethod{} to it, replacing depth sensors with VGGT to construct the pointcloud. We follow a similar protocol as~\ref{sec:Viewpoint_Variance_Evaluation} using a dataset with diverse viewpoints, $D_{diverse}$, for training and then evaluating the policies in 5 novel camera configurations that span the table.

We train both versions of 3D Diffuser Actor with 24000 training steps using $D_{diverse}$ as the training dataset. Since 3D diffuser actor actually interacts directly with a point cloud, we do not obtain a canonical viewpoint when training the policy with \ourmethod{}. 
\begin{wrapfigure}{r}{0.5\textwidth}
    \vspace{-2mm}
    \centering
        \includegraphics[width=0.5\textwidth]{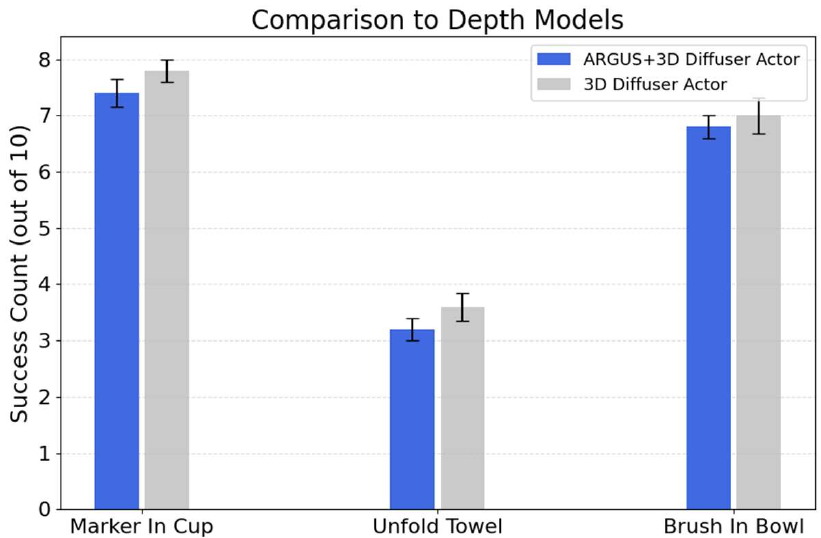}
    \caption{Success rate comparison between our method and a depth-based policy baseline.}
    \label{fig:depth_results}
    \vspace{-20mm}
\end{wrapfigure}Instead, we feed the scaled VGGT-predicted 3D pointcloud directly into 3D Diffuser Actor's transformer architecture.

\subsubsection{Results}
We report the average success rate of both classical 3D Diffuser Actor and \ourmethod{}+3D Diffuser Actor (without depth sensor) in Figure~\ref{fig:depth_results}. As seen in Figure~\ref{fig:depth_results}, using VGGT to construct the pointcloud performs comparably with using raw depth data. 

\section{Limitations}
\label{sec:limitations}
While \ourmethod{} demonstrates strong viewpoint generalization across the manipulation tasks considered in this work, it exhibits limitations in tasks requiring high spatial precision. In particular, we observe degraded performance on fine-grained manipulation tasks involving small objects (e.g., grasping a small button). When further analyzing this failure mode, we observe two primary causes. First, we observe the scene experiencing slight translational shifts between consecutive canonical images. We hypothesize that these shifts arise from small extrinsic prediction errors by VGGT that compound during canonical frame alignment as seen in Section~\ref{sec:alignment_sec}. Such variation can introduce minor inconsistencies in the visual observation space, ultimately causing the downstream policy to predict slightly misaligned actions. Second, we found that occlusion-caused holes in the reconstructed point cloud and canonical frame rendering may introduce artifacts that occasionally drown out small objects. Both these observations lead to the conclusion that \ourmethod{} currently trades some degree of spatial quality and precision for viewpoint robustness. 

A second inherent limitation within \ourmethod{} is the added computational overhead incurred at evaluation time. Because we run VGGT at every policy prediction, \ourmethod{} introduces an average latency of about 0.52s per prediction using an RTX 3080 GPU. While acceptable for most quasistatic tasks like the ones studied in this paper, this overhead may limit the applicability of \ourmethod{} in tasks requiring immediate closed-loop responses or high-frequency control.
\section{Conclusion}
\label{sec:conclusion}
In this paper, we presented \ourmethod{}, a simple yet effective image preprocessing pipeline that enables visuomotor policies to learn efficiently from datasets regardless of the degree of viewpoint variation present. By leveraging pretrained 3D vision models and their learned geometric priors, \ourmethod{} transforms observations captured from arbitrary viewpoints into normalized canonical views, reducing viewpoint-induced variability before policy learning. This decomposition allows downstream visuomotor policies to focus on task-relevant structure rather than disentangling camera pose from scene understanding. Through extensive evaluation across large viewpoint variations spanning the scene, we demonstrate that \ourmethod{} improves both learning efficiency and zero-shot generalization to novel viewpoints compared to baseline approaches. These results suggest that explicitly canonicalizing observations provides a practical and scalable alternative to learning viewpoint invariance solely through large-scale data collection. Future work may extend this framework beyond viewpoint normalization toward broader forms of scene invariance, as well as improve geometric fidelity to better support fine-grained manipulation tasks while maintaining viewpoint robustness.

%===============================================================================

% The acknowledgments are automatically included only in the final and preprint versions of the paper.
\acknowledgments{We thank Kaiwen Hong for assistance with hardware setup. This work used the Delta system at the National Center for Supercomputing Applications [award OAC 2005572] through allocation CIS260352 from the Advanced Cyberinfrastructure Coordination Ecosystem: Services \& Support (ACCESS) program, which is supported by National Science Foundation grants \#2138259, \#2138286, \#2138307, \#2137603, and \#2138296.}

%===============================================================================

\bibliography{references}
\clearpage
\appendix

\section*{Appendix}
% \addcontentsline{toc}{section}{Appendix}

% \startcontents[appendix]
% \printcontents[appendix]{l}{1}{}
% \newpage
\section{Additional Experimental Details and Results}
\subsection{Hardware Details}
Our experimental setup can be visualized in Figure~\ref{fig:setup}. It includes two \textbf{Intel Realsense D415 cameras} and four \textbf{April Tags}~\cite{apriltags} which are used to quickly calibrate each camera at the start of each demo. We use a 7-dof \textbf{Kinova Gen 3} robot arm along with a \textbf{robotiq hand-e}.
\subsection{Additional High/low Viepoint Variance Evaluation Details}

\begin{table*}[!b]
\centering
\caption{Success counts (out of 10) across five camera positions. Mean $\pm$ standard error are reported over the five viewpoints.}

\label{tab:camera_generalization}
\renewcommand{\arraystretch}{1.2}
\begin{tabular}{llccccc@{\hskip 0.8em}|@{\hskip 0.8em}c}
\toprule
\textbf{Task} & \textbf{Method} &
\textbf{Cam (a)} & \textbf{Cam (b)} & \textbf{Cam (c)} &
\textbf{Cam (d)} & \textbf{Cam (e)} &
\textbf{Mean $\pm$ Std. Error} \\
\midrule

\multirow{3}{*}{\begin{tabular}[c]{@{}l@{}}Marker\\in Cup\end{tabular}}
& \ourmethod{} & \textbf{6} & \textbf{6} & \textbf{4} & \textbf{6} & \textbf{5} & \textbf{5.4 $\pm$ 0.400} \\
& DP           & 1 & 1 & 2 & 0 & 1 & 1.0 $\pm$ 0.316 \\
& KYC          & 0 & 0 & 0 & 0 & 0 & 0.0 $\pm$ 0.000 \\
\midrule

\multirow{3}{*}{\begin{tabular}[c]{@{}l@{}}Unfold\\Towel\end{tabular}}
& \ourmethod{} & \textbf{9} & \textbf{7} & \textbf{9} & 7 & \textbf{8} & \textbf{8.0 $\pm$ 0.447} \\
& DP           & 5 & \textbf{7} & \textbf{8} & 6 & 2 & 5.6 $\pm$ 1.030 \\
& KYC          & 8 & 6 & 1 & \textbf{8} & 2 & 5.0 $\pm$ 1.483 \\
\midrule

\multirow{3}{*}{\begin{tabular}[c]{@{}l@{}}Put Brush\\in Bowl\end{tabular}}
& \ourmethod{} & \textbf{5} & \textbf{6} & \textbf{6} & \textbf{5} & \textbf{4} & \textbf{5.2 $\pm$ 0.374} \\
& DP           & 0 & 1 & 2 & 1 & 0 & 0.8 $\pm$ 0.374 \\
& KYC          & 1 & 2 & 2 & 0 & 0 & 1.0 $\pm$ 0.447 \\
\midrule

\multirow{3}{*}{\begin{tabular}[c]{@{}l@{}}Stack\\Blocks\end{tabular}}
& \ourmethod{} & \textbf{9} & \textbf{9} & \textbf{8} & \textbf{7} & \textbf{7} & \textbf{8.0 $\pm$ 0.447} \\
& DP           & 0 & 1 & 1 & 1 & 1 & 0.8 $\pm$ 0.200 \\
& KYC          & 0 & 0 & 0 & 0 & 0 & 0.0 $\pm$ 0.000 \\
\bottomrule
\end{tabular}
\end{table*}

\begin{table*}[!b]
\centering
\caption{Success counts (out of 10) across five camera positions. Mean $\pm$ standard error are reported over the five viewpoints.}
\label{tab:camera_vista_generalization}
\renewcommand{\arraystretch}{1.2}
\begin{tabular}{llccccc@{\hskip 0.8em}|@{\hskip 0.8em}c}
\toprule
\textbf{Task} & \textbf{Method} &
\textbf{Cam (a)} & \textbf{Cam (b)} & \textbf{Cam (c)} &
\textbf{Cam (d)} & \textbf{Cam (e)} &
\textbf{Mean $\pm$ Std. Error} \\
\midrule

\multirow{3}{*}{\begin{tabular}[c]{@{}l@{}}Marker\\in Cup\end{tabular}}
& \ourmethod{} & \textbf{8} & \textbf{8} & \textbf{5} & \textbf{8} & \textbf{6} & \textbf{7.0 $\pm$ 0.632} \\
& DP           & 5 & 0 & 0 & 0 & 0 & 1.0 $\pm$ 1.000 \\
& VISTA        & 1 & 0 & 1 & 0 & 0 & 0.4 $\pm$ 0.245 \\
\midrule

\multirow{3}{*}{\begin{tabular}[c]{@{}l@{}}Unfold\\Towel\end{tabular}}
& \ourmethod{} & \textbf{10} & \textbf{10} & \textbf{8} & \textbf{9} & \textbf{8} & \textbf{9.0 $\pm$ 0.447} \\
& DP           & 8 & 2 & 0 & 0 & 0 & 2.0 $\pm$ 1.549 \\
& VISTA        & 6 & 6 & 3 & 5 & 5 & 5.0 $\pm$ 0.548 \\
\midrule

\multirow{3}{*}{\begin{tabular}[c]{@{}l@{}}Put Brush\\in Bowl\end{tabular}}
& \ourmethod{} & \textbf{7} & \textbf{7} & \textbf{4} & \textbf{5} & \textbf{4} & \textbf{5.4 $\pm$ 0.678} \\
& DP           & 6 & 2 & 0 & 0 & 0 & 1.6 $\pm$ 1.166 \\
& VISTA        & 1 & 2 & 0 & 0 & 1 & 0.8 $\pm$ 0.374 \\
\midrule

\multirow{3}{*}{\begin{tabular}[c]{@{}l@{}}Stack\\Blocks\end{tabular}}
& \ourmethod{} & \textbf{10} & \textbf{8} & \textbf{8} & \textbf{5} & \textbf{6} & \textbf{7.4 $\pm$ 0.872} \\
& DP           & 4 & 1 & 0 & 0 & 0 & 1.0 $\pm$ 0.775 \\
& VISTA        & 0 & 1 & 1 & 0 & 0 & 0.4 $\pm$ 0.245 \\
\bottomrule

\end{tabular}
\end{table*}
We conduct our main evaluation study as initially discussed in Section~\ref{sec:Viewpoint_Variance_Evaluation} by dividing it into two trial dimensions: object positions and camera positions. We sample 10 random object configurations within the task space and group them into one set of trials. We then repeat these set of trials for 5 different camera viewpoint configurations as discussed in Section~\ref{sec:Viewpoint_Variance_Evaluation} and visualized in Figure~\ref{fig:eval-cams}. We report the numeric results from each set of trials in Table~\ref{tab:camera_generalization} and Table~\ref{tab:camera_vista_generalization}. The mean and standard errors reported are computed over the 5 different sets of trials and are visualized in Figure~\ref{fig:viewpoint_variance_eval}

\end{document}